\documentclass{article}

\usepackage{PRIMEarxiv}

\usepackage[utf8]{inputenc} 
\usepackage[T1]{fontenc}    
\usepackage{hyperref}       
\usepackage{url}            
\usepackage{booktabs}       
\usepackage{amsfonts}       
\usepackage{nicefrac}       
\usepackage{microtype}      
\usepackage{lipsum}
\usepackage{fancyhdr}       
\usepackage{graphicx}       
\graphicspath{{media/}}     

\usepackage[T1]{fontenc}
\usepackage{graphicx}
\usepackage{booktabs}
\usepackage[misc]{ifsym}

\usepackage{hyperref}
\usepackage{caption}
\usepackage{subcaption}
\usepackage{amsmath}
\usepackage{amssymb}
\usepackage{multicol}
\usepackage{multirow}
\usepackage{caption}
\usepackage{subcaption}
\usepackage{float}
\usepackage{tikz}
\usepackage{ctable}
\usepackage{bm}
\usepackage{algorithm}
\usepackage{wrapfig}
\usepackage{algpseudocodex}
\algrenewcommand{\algorithmiccomment}[1]{\hfill\textbf{//}\,#1}
\usepackage{mwe}
\usepackage{cite}
\def \knn {FeNeC}
\def \logit {FeNeC-Log}

\DeclareMathOperator{\argmax}{arg\,max}

\title{FeNeC: Enhancing Continual Learning via Feature Clustering with Neighbor- or Logit-Based Classification}

\author{
  Kamil Książek, Hubert Jastrzębski$^{*}$, Jacek Tabor \\
  Faculty of Mathematics and Computer Science \\
  Jagiellonian University \\
  \texttt{kamil.ksiazek@uj.edu.pl} \\ 
  \texttt{h.jastrzebski@student.uj.edu.pl} \\
  \texttt{jacek.tabor@uj.edu.pl} \\
  \And
  Krzysztof Pniaczek$^{*}$ \\
  Secondary School No 2 in Nowy Targ \\
  \texttt{krzysiek.pniaczek@outlook.com}
  \And
  Bartosz Trojan$^{*}$, Michał Karp$^{*}$ \\
  Upper-Secondary Schools of Communications in Cracow \\
 \texttt{bartosztrojanofficial@gmail.com} \\
 \texttt{contact@michalkarp.pl}
}

\begin{document}
\maketitle

\maketitle              

\begin{abstract}
The ability of deep learning models to learn continuously is essential for adapting to new data categories and evolving data distributions. In recent years, approaches leveraging frozen feature extractors after an initial learning phase have been extensively studied. Many of these methods estimate per-class covariance matrices and prototypes based on backbone-derived feature representations. Within this paradigm, we introduce FeNeC (Feature Neighborhood Classifier) and FeNeC-Log, its variant based on the log-likelihood function. Our approach generalizes the existing concept by incorporating data clustering to capture greater intra-class variability. Utilizing the Mahalanobis distance, our models classify samples either through a nearest neighbor approach or trainable logit values assigned to consecutive classes. Our proposition may be reduced to the existing approaches in a special case while extending them with the ability of more flexible adaptation to data. We demonstrate that two FeNeC variants achieve competitive performance in scenarios where task identities are unknown and establish state-of-the-art results on several benchmarks.

\end{abstract}

\def\thefootnote{*}\footnotetext{These authors contributed equally to this work}\def\thefootnote{\arabic{footnote}}

\section{Introduction}


The ability to acquire new information incrementally, rather than all at once, is a critical skill exhibited by living organisms. Humans, for example, continuously learn throughout their lives, in contrast to traditional machine learning algorithms, which typically achieve optimal performance only when trained on all available data at once~\cite{kirkpatric2017OCF}. In deep learning \cite{schmidhuber2015DL}, however, it is common to encounter scenarios where data arrives sequentially, leading to a significant challenge for standard neural networks: the phenomenon of catastrophic forgetting \cite{delange2019CL, kirkpatric2017OCF, mccloskey1989CIiCN}, where previously learned knowledge is overwritten or lost. This challenge is the focus of continual learning (CL), also referred to as lifelong learning. The goal of CL is to enable machine learning models to learn new tasks over time while preserving knowledge from previously encountered tasks \cite{wang2024CSoCL, verwimp2024CLApplications}.

A main challenge in continual learning is balancing the stability-plasticity trade-off \cite{wang2024CSoCL, abraham2005MR}. Stability refers to the model's ability to retain prior knowledge, while plasticity is its capacity to adapt to new data. Excessive stability can hinder the learning of new information, while excessive plasticity can lead to forgetting previously acquired knowledge.

Various strategies have been proposed to address this issue. One of the possible solutions could be storing examples from prior tasks for later use, but this can pose challenges related to storage limitations and privacy concerns. Additionally, iterative retraining to incorporate new data increases computational demands and processing time. In the specific setting of exemplar-free Class-Incremental Learning (CIL), models are designed to learn from data that arrives incrementally, with disjoint label spaces between tasks \cite{zhou2023RevisitingCIL, hsu2018RCLS, masana2023CIL}. Notably, task identities are provided only during training, further complicating inference.

Backbone freezing methods have recently gained popularity in this setting. In visual classification, a common approach pre-trains the model backbone on the initial task and subsequently freezes the feature extractor \cite{ma2023progressive, petit2023fetril, goswami2023fecam}, leveraging its ability to generalize across distributions. Following \cite{petit2023fetril, goswami2023fecam}, class centroids are computed and stored for classification. We generalize the FeCAM approach \cite{goswami2023fecam} and propose \knn{} and its variant \logit{}, both relying on a pre-trained, frozen backbone. \knn{} stores $N_{clusters}$ centroids per class and classifies new samples via a Mahalanobis distance-based weighted k-Nearest Neighbors (kNN) approach \cite{cunningham2022KNN}. \logit{} introduces two trainable parameters shared across classes, computing log-likelihoods over neighboring class points to determine predictions. We present our approach in detail and evaluate it on standard continual learning benchmarks, achieving state-of-the-art performance in several settings.

We can summarize our contributions in the following way:
\begin{itemize}
    \item We propose \knn{} and its variant, \logit{}, which leverage data clustering to model intra-class variability and perform classification based on neighboring points or the log-likelihood function.
    \item Our approach extends FeCAM which represents each class with a single prototype. Instead, we utilize multiple centroids. Notably, in the trivial case, \knn{} may be reduced to FeCAM.
    \item We demonstrate state-of-the-art performance across several continual learning benchmarks.
\end{itemize}

\begin{figure}[!t]
    \centering
    \includegraphics[trim={0cm, 0cm, 0cm, 0cm},clip,width=0.85\linewidth]{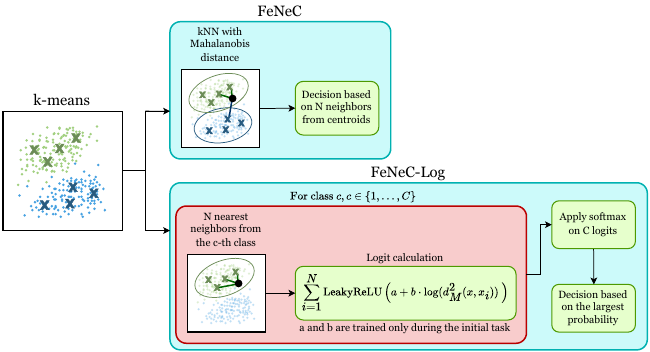}%
    \captionof{figure}{The workflow of the methods presented in this paper. Initially, $k$--means clustering is performed on features extracted from the neural network. Then, in~\knn, a kNN classifier with Mahalanobis distance is used, while in~\logit, each class has a corresponding logit value with shared parameters $a, b \in \mathbb{R}$, trained only during the initial task.\label{fig:teaser}}
\end{figure}

\section{Related Works}


Continuous learning without forgetting previous data distributions is not a trivial issue. As a result, many completely different approaches have been proposed over the years. Among these methods we can distinguish replay-based, architecture-based, regularization-based or optimization-based (gradient-based)~\cite{wang2024CSoCL}.

\underline{Replay-based} is a group of methods that focus on replaying exemplars from the data distribution of previous tasks while learning new classes \cite{wang2024CSoCL}. In practical applications, there are legal issues related to data storage, so methods such as FSCIL via Entropy-Regularized Data-Free Replay, Deep Generative Replay, and EEC utilize generative networks that mimic the distribution of previous tasks \cite{liu2022few, shin2017continual, ayub2021eec}. However, it has been observed that the quality of synthetic data degrades as the number of classes increases, so methods such as Hybrid Memory Replay or CloGAN combine synthetic data with carefully selected real examples \cite{kong2024hybrid, rios2018closed}.

\underline{Architecture-based} methods try to design architectures with task-specific parameters \cite{wang2024CSoCL}. For instance, PackNet performs aggressive parameter pruning and retrains the network, making it possible to learn multiple tasks through a single model \cite{mallya2018packnet}. Other approaches, WSN and SoftNet, inspired by the Lottery Ticket Hypothesis (LTH), create parameter masks based on their importance \cite{kang2022WSN, frankle2018lth}. The other approach, UCL, combines the KL-divergence term, the Gaussian mean-field approximation and the Bayesian neural network pruning \cite{ahn2019UCL}. 

\underline{Optimization-based} methods utilize the properties of the optimization process to continuously learn new tasks~\cite{wang2024CSoCL}. An exemplary approach is the GEM method, which uses gradient projection to ensure that gradients from the current task do not interfere with previously learned tasks~\cite{lopez2017GEM}. 
A-GEM averages the imposed constraints \cite{chaudhry2018agem} while LOGD decomposes the gradient into a joint component and a class-specific component for each layer separately \cite{tang2021layerwise}. Another method is TRGP, which uses gradients to identify previous tasks strongly correlated with the current one~\cite{lin2022trgp}. Some recent works consider hypernetworks: methods like HNET~\cite{oswald2019hypercl}, HyperMask~\cite{książek2023hypermask}, and Partial Hypernetworks~\cite{hemati2023partial} optimize task-specific embeddings or masks to dynamically regulate parameter updates, balancing plasticity and stability through lightweight architectural adaptations. Posterior Meta-Replay\cite{henning2021posterior} combines Bayesian inference with hypernetworks.

\underline{Regularization-based} aim to minimize modifications to weights crucial for previous tasks. For example, EWC~\cite{kirkpatric2017OCF} uses Fisher information matrix (FIM) to identify important parameters, whereas SI~\cite{zenke2017SI} and MAS~\cite{aljundi2018MAS} estimate their importance through a weight’s contribution to global loss and prediction sensitivity to parameter changes, respectively. In contrast, AFEC~\cite{wang2021afec} adopts an approach inspired by biological active forgetting, selectively forgetting knowledge that interferes with new tasks. 


Recently, methods using \underline{frozen backbones} have become a subject of intense research. 
The authors in~\cite{janson2023simplebaseline} use pre-trained models and the Nearest Mean Classifier, calculating mean class representations and their Euclidean distance to test samples. FeTrIL~\cite{petit2023fetril} reintroduces a linear classifier but freezes representations and reconstructs past distributions using class prototypes. FeCAM~\cite{goswami2023fecam} applies frozen backbones with an NCM classifier incorporating covariance and Mahalanobis distance. H-Prompts~\cite{gao2024prompts} employs learnable class, task and general prompts to pre-trained transformers. AdaGauss~\cite{rypesc2025task} unfreezes representations while leveraging covariance. These studies highlight a shift toward strong feature extraction, geometric vector interpretation, and class-specific covariances over linear classification.


\section{Methods}

\begin{algorithm}[!t]
	\caption{The pseudocode of \knn{} algorithm.\label{alg:fenec}}
	\begin{algorithmic}[1]
        \Require $T$, $\mathcal{D}_{1}^{train}, ..., \mathcal{D}_{T}^{train}$, $\mathcal{D}_{1}^{test}, ..., \mathcal{D}_{T}^{test}$, 
		$N_{clusters}$, $N_{neighbors}$, $N_{classes}$, $C$.
		\Ensure class prediction for all test samples from $T$ tasks.
		\For{task $t \leftarrow 1$ to $T$}
			\For{class $c \in C_t$}
				\State Select all training samples belonging to the $c$--th class, $(\mathbf{X_c}, \mathbf{Y_c}) \subseteq \mathcal{D}_{t}^{train}$.
                \State Apply Tukey's transformation on $\mathbf{X_c}$, creating $T(\mathbf{X_c})$.
                \State Create a covariance matrix for the $c$--th class based on $T(\mathbf{X_c})$, apply shrinkage and normalization, creating $\mathbf{\Sigma_{c}}$.
				\State Apply the k-means clustering and select $N_{clusters}$ representatives of the $c$--th class samples, $R_c = \lbrace \mathcal{X}_c^{1}, \mathcal{X}_c^{2}, ..., \mathcal{X}_c^{N_{clusters}} \rbrace$.
			\EndFor
			\For{$(\mathbf{x}, y) \in \mathcal{D}_{t}^{test}$}
                \State Apply Tukey's transformation on $\mathbf{x}$, creating $T(\mathbf{x})$.
				\State Calculate distances between $T(\mathbf{x})$ and all individual classes' representatives from the set $R = R_1 \cup R_2 \cup ... \cup R_t$, in terms of the Mahalanobis distance.
				\State Select the class for $T(\mathbf{x})$ using the kNN classifier, $N_{neighbors}$ nearest representatives and a weighted distance rule.
			\EndFor
		\EndFor
	\end{algorithmic}
\end{algorithm}

Let us suppose that we have $T$ continual learning tasks. Each task $t$, where $t \in \lbrace 1, ..., T \rbrace$, has its own training set $\mathcal{D}_{t}^{train}$, containing $N_t^{train}$ data samples and corresponding labels, i.e. $\mathcal{D}_{t}^{train} = \left\{ (\mathbf{x}^{train}_{j, t}, y^{train}_{j, t})~|~j \in \lbrace 1, ..., N_t^{train} \rbrace \right\}$. Let $C$ be the set containing unique classes being present in $T$ tasks while $C_t$ will be the set of unique classes for samples from the $t$--th task and $N_{t}^{classes} = |C_t|$. Therefore, $C = C_1 \cup C_2 \cup ... \cup C_T$, and the total number of classes for the entire dataset is equal to $N_{classes} = |C|$.

We consider only the Class-Incremental Learning scenario in which task identity is known during training, and not in inference~\cite{Vandeven2022scenarios}. The test set for the $t$-th task consists of $N_t^{test}$ samples, $\mathcal{D}_{t}^{test} = \left\{ \mathbf{x}_{j, t}^{test} ~|~ j \in \lbrace 1, ..., N_t^{test} \rbrace \right\}$, and may store exemplars from classes present both in the $t$--th and the preceding tasks. Therefore, indicating the correct class becomes more difficult as $t$ increases.

In this paper, we present \textbf{\knn} (Feature Neighborhood Classifier) and its variant \textbf{\logit} (Feature Neighborhood Classifier with Log-Likelihood). They draw inspiration from FeCAM, i.e. the solution presented in~\cite{goswami2023fecam}. \knn{} and \logit{} extend this approach, beating the baseline in some of the considered experimental scenarios. The authors of FeCAM proposed a calculation of the covariance matrix and mean vector per each class, based on the features extracted from the backbone network, frozen after the initial task. Finally, for each new sample, the Mahalanobis distance between this point and the mean vectors of consecutive classes was calculated.

However, while one covariance matrix per class is more robust than a single covariance matrix for all data samples, we claim that it still may not represent the intra-class data variability. Therefore, in the first stage, we propose k--means clustering, to distinguish several groups of points within a given class. In the next phase, we apply two different approaches. In \knn, we apply a k--nearest neighbor (kNN) classifier which is an efficient non-parametric algorithm. Instead of original data points, we calculate the distance between a sample and centroids representing subsequent classes. Also, we use the Mahalanobis distance in the kNN classifier, instead of the typical Euclidean one.  

Both methods start on the output of an external feature extractor like Visual Transformer or ResNet. Then, the consecutive steps of the presented methods are, as follows.

\subsection{\knn{}}\label{sec:fenec}
Let us assume that we consider the $t$--th task, where $t \in \lbrace 1, ..., T \rbrace$. We start our algorithm with the separation of all samples assigned to the $c$--th class, where $c \in \lbrace 1, ..., N_{t}^{classes} \rbrace$, and we call it by $\mathbf{X_c}$. Similarly as e.g. in FeCAM~\cite{goswami2023fecam}, for each class $c$, we calculate a covariance matrix $\mathbf{\hat{\Sigma}_c}$. However, in the initial step, in cases, where all returned values are non-negative, we perform Tukey's transformation on extracted features, i.e. we calculate $T(\mathbf{X_c}) = \mathbf{X_c}^{\lambda}$, where $\lambda > 0$ is a hyperparameter. The covariance matrix is estimated based on those reshaped features.

Due to the possibility of having fewer samples than the number of matrix dimensions, to obtain a more advantageous and full-rank covariance matrix representation, we apply covariance shrinkage and normalization. Firstly, the shrunk covariance matrix $\mathbf{\hat{\Sigma}^{s}_c}$ is created. To all elements located in the main diagonal of the estimated covariance matrix $\mathbf{\hat{\Sigma}_c}$, the average of all diagonal elements multiplied by the $\gamma_1$ value is added. For the remaining elements, the average of all matrix elements located outside of the main diagonal, multiplied by the $\gamma_2$ value, is added. Both $\gamma_1$ and $\gamma_2$ are hyperparameters. 

Let us suppose that $\mathbf{\hat{\Sigma}_c} \in \mathbb{R}^{N_{f} \times N_{f}}$, where $N_f$ is the number of features, i.e. the size of the feature extractor output. Then
\begin{equation}
    \mathbf{\hat{\Sigma}^{s}_c} = \mathbf{\hat{\Sigma}_c} + \gamma_{1} \cdot \frac{1}{N_{f}} \sum\limits_{i=1}^{N_{f}}{\mathbf{\hat{\Sigma}_c}(i, i)} \cdot \mathbb{I} + \gamma_{2} \cdot \frac{1}{N^2_{f} - N_{f}} \sum_{\substack{i, j = 1 \\ i \neq j}}^{N_{f}} {\mathbf{\hat{\Sigma}_c}(i, j)} \cdot (1 - \mathbb{I}),
\end{equation}
\noindent where $\mathbb{I} \in \mathbb{R}^{N_{f} \times N_{f}}$ is an identity matrix and ${\mathbf{\hat{\Sigma}_c}(i, j)}$ denotes the $(i, j)$--th position of $\mathbf{\hat{\Sigma}_c}$. In the next step, to reduce differences in variances between classes, we normalize the shrunk covariance matrix, creating its final representation
\begin{equation}
    \mathbf{\Sigma_c} = \dfrac{\mathbf{\hat{\Sigma}^{s}_c}(i, j)}{\sqrt{\mathbf{\hat{\Sigma}^{s}_c}(i, i) \cdot \mathbf{\hat{\Sigma}^{s}_c}(j, j)}}.
\end{equation}

\begin{algorithm}[!t]
	\caption{The pseudocode of \logit{} algorithm.\label{alg:feloc}}
	\begin{algorithmic}[1]
        \Require $T$, $\mathcal{D}_{1}^{train}, ..., \mathcal{D}_{T}^{train}$, $\mathcal{D}_{1}^{test}, ..., \mathcal{D}_{T}^{test}$,
		$N_{clusters}$, $N_{points}$, $N_{classes}$, $C$, $\hat{C_t} =  C_1 \cup C_2 \cup ... \cup C_t$: the set containing unique classes that appeared up to the $t$--th task, $N_{classes}^{:t} = |\hat{C_t}|$.
		\Ensure class prediction for all test samples from $T$ tasks.
		\State Initialize trainable $a, b \in \mathbb{R}$.
		\For{task $t \leftarrow 1$ to $T$}
			\For{class $c \in C_t$}
				\State Select all training samples belonging to the $c$--th class, $(\mathbf{X_c}, \mathbf{Y_c}) \subseteq \mathcal{D}_{t}^{train}$.
                \State Apply Tukey's transformation on $\mathbf{X_c}$, creating $T(\mathbf{X_c})$.
                \State Create a covariance matrix for the $c$--th class based on $T(\mathbf{X_c})$, apply shrinkage and normalization, creating $\mathbf{\Sigma_{c}}$.
				\State Apply the k-means clustering and select $N_{clusters}$ representatives of the $c$--th class samples, $R_c = \lbrace \mathcal{X}_c^{1}, \mathcal{X}_c^{2}, ..., \mathcal{X}_c^{N_{clusters}} \rbrace$.
            \EndFor
                \If{t == 1}
                    \For{$(\mathbf{x}, y) \in \mathcal{D}_{1}^{train}$}
                        \State Apply Tukey's transformation on $\mathbf{x}$, creating $T(\mathbf{x})$.
                        \For{class $c \in \hat{C_1}$}
                            \State Select $\overline{R}_{c, T(\mathbf{x})}$ containing $N_{points}$ nearest $c$--th class centroids for $T(\mathbf{x})$, i.e. $\overline{R}_{c, T(\mathbf{x})} = \lbrace \overline{{\mathcal{X}}}_{c}^{1}, \overline{{\mathcal{X}}}_{c}^{2}, ..., \overline{{\mathcal{X}}}_{c}^{N_{points}} \rbrace$.
                            \State $logit_{c}(T(\mathbf{x})) = \sum\nolimits_{j=1}^{N_{points}} \mathrm{LeakyReLU} \bigg( a + b \cdot \log \big( d_M^2( T(\mathbf{x}), \overline{{\mathcal{X}}}_{c}^{j} \big) \big) \bigg)$, \\
                            $d_M^2( T(\mathbf{x}), \overline{{\mathcal{X}}}_{c}^{j} \big) = (T(\mathbf{x}) - \overline{{\mathcal{X}}}_{c}^{j})^\top  \mathbf{\Sigma^{-1}_{c}} (T(\mathbf{x}) - \overline{{\mathcal{X}}}_{c}^{j})$
                        \EndFor
                        \State Apply softmax to all logit values.
                        \State Train $a$ and $b$, maximizing a logit value for the ground truth class.
                    \EndFor
                \EndIf
			\For{$(\mathbf{x}, y) \in \mathcal{D}_{t}^{test}$}
                \State Apply Tukey's transformation on $\mathbf{x}$, creating $T(\mathbf{x})$.
                \For{class $c \in \hat{C_t}$}
                    \State Select $\overline{R}_{c, T(\mathbf{x})}$ containing $N_{points}$ nearest $c$--th class centroids for $T(\mathbf{x})$, i.e. $\overline{R}_{c, T(\mathbf{x})} = \lbrace \overline{{\mathcal{X}}}_{c}^{1}, \overline{{\mathcal{X}}}_{c}^{2}, ..., \overline{{\mathcal{X}}}_{c}^{N_{points}} \rbrace$.
                    \State $logit_{c}(T(\mathbf{x})) = \sum\nolimits_{i=1}^{N_{points}} \mathrm{LeakyReLU} \left( a + b \cdot \log \left( d_M^2( T(\mathbf{x}), \overline{{\mathcal{X}}}_{c}^{i}) \right) \right)$
                \EndFor
                \State Apply softmax to all logits, creating $\overline{logit}_{c}(T(\mathbf{x})$, $c \in \lbrace 1, ..., N_{classes}^{:t} \rbrace$.
                \State Select the largest logit's class: $\hat{y} = \argmax\limits_{c \in \hat{C_t}} \overline{logit}_{c}(T(\mathbf{x}))$.
			\EndFor
		\EndFor
	\end{algorithmic}
\end{algorithm}

After estimation of the covariance matrix for the $c$--th class, we want to better express the intra-class diversity than through a FeCAM-like single prototype calculation. Therefore, we use $k$--means clustering to select $N_{clusters}$ centroids for a given class, $R_c = \lbrace \mathcal{X}_c^{1}, \mathcal{X}_c^{2}, ..., \mathcal{X}_c^{N_{clusters}} \rbrace$, where $\mathcal{X}_c^{j} \in \mathbb{R}^{N_{f}}$ is the $j$--th centroid coordinates for the $c$--th class,   $N_{clusters}$ is a hyperparameter. 

During inference in the $t$--th task, instead of a selection of the nearest prototype like in FeCAM, we use a weighted $k$--nearest neighbor classifier with Mahalanobis distance. Assume that $N_{classes}^{:t}$ is the number of classes from all tasks up to task $t$. We want to assign a label to the Tukey-transformed sample $\mathbf{x} \in \mathbb{R}^{N_f}$. In this goal, we create a set of all centroids created so far $R = \lbrace \mathcal{X}_1^{1}, ..., \mathcal{X}_1^{N_{clusters}}, ..., \mathcal{X}_{N_{classes}^{:t}}^{1}, ..., \mathcal{X}_{N_{classes}^{:t}}^{N_{clusters}} \rbrace$.  We calculate the squared Mahalanobis distance between those centroids and $\mathbf{x}$, i.e. $d_M^2( \mathbf{x}, \mathcal{X}_{c}^{j} \big) = (\mathbf{x} - \mathcal{X}_{c}^{j})^\top  \mathbf{\Sigma^{-1}_{c}} (\mathbf{x} - \mathcal{X}_{c}^{j})$, where $\mathcal{X}_{c}^{j}$ is the $j$--th centroid of the $c$--th class. Then, we select $N_{neighbors}$ lowest distances, $N_{neighbors} \leq N_{classes}*N_{clusters}$, and we calculate the value of $d_{kNN}$ between $\mathbf{x}$ and its nearest neighbors belonging to particular classes (if any), where $N_{neighbors} = N_{neighbors}^{1} + N_{neighbors}^{2} + ... + N_{neighbors}^{N_{classes}^{:t}}$. For a given class $c$ and sample $\mathbf{x}$
\begin{equation}
    d_{kNN}(\mathbf{x}, c) = \sum\limits_{j=1}^{N_{neighbors}^{c}} \frac{1}{d_M^2( \mathbf{x}, \overline{\mathcal{X}}_{c}^{j} )},
\end{equation}
where $\overline{\mathcal{X}}_{c}^{j}$ is the $j$--th nearest cluster representing the $c$--th class, $j \in \lbrace 1, ..., N_{neighbors}^{c} \rbrace$. Finally, we assign the class with the largest value of $d_{kNN}$, i.e. $\hat{y} = \argmax_{j \in \lbrace 1, ..., N_{classes}^{:t} \rbrace} d_{kNN}(\mathbf{x}, j)$. The pseudocode of this approach is presented in Algorithm~\ref{alg:fenec}. \knn{} needs storing $N_{classes} \cdot N_{f}^{2}$ parameters for covariance matrices and $N_{classes} \cdot N_{clusters} \cdot N_{f}$ parameters for centroids. In total, it requires $N_{classes} \cdot N_{f} (N_{clusters} + N_{f})$ parameters in memory.

\subsection{\logit{}}
The initial steps of the \logit{} algorithm, including applying the Tukey transformation to features, estimating the covariance matrix, its shrinkage and normalization, and $k$--means clustering, are the same as in \knn{}. However, the decision about sample labels is made differently.

Let us consider a sample $\mathbf{x} \in T(\mathbf{X_c})$.  During the first task, for every class $c \in \lbrace 1, ..., N_{classes}^{1} \rbrace$ that appears in this task, we calculate a dedicated logit with shared parameters $a, b \in \mathbb{R}$. For a given point $\mathbf{x}$, we start from identifying $N_{points}$ nearest centroids of the $c$--th class, $\overline{R}_{c, \mathbf{x}} = \lbrace \overline{{\mathcal{X}}}_{c}^{1}, \overline{{\mathcal{X}}}_{c}^{2}, ..., \overline{{\mathcal{X}}}_{c}^{N_{points}} \rbrace$, according to the Mahalanobis distance. Then, we extend the $k$--nearest neighbors approach, through the calculation of a log-likelihood function with trainable parameters
\begin{equation}
    logit_{c}(\mathbf{x}) = \sum\nolimits_{i=1}^{N_{points}} \mathrm{LeakyReLU} \left( a + b \cdot \log \left( d_M^2( \mathbf{x}, \overline{{\mathcal{X}}}_{c}^{i}) \right) \right),
\end{equation}
\noindent where $d_M^2( \mathbf{x}, \overline{{\mathcal{X}}}_{c}^{i})$ is the squared Mahalanobis distance between $\mathbf{x}$ and the $i$--th nearest centroid, i.e. $d_M^2( \mathbf{x}, \overline{{\mathcal{X}}}_{c}^{i} \big) = (\mathbf{x} - \overline{{\mathcal{X}}}_{c}^{i})^\top  \mathbf{\Sigma^{-1}_{c}} (\mathbf{x} - \overline{{\mathcal{X}}}_{c}^{i})$.
Next, we apply softmax to all logit values:
\begin{equation}
    \overline{logit}_c(\mathbf{x}) = \frac{\exp \left( logit_c(\mathbf{x}) \right)}{\sum\nolimits_{j=1}^{N_{classes}^{1}} \exp(logit_j(\mathbf{x}))}.
\end{equation}
Finally, we predict the label for $\mathbf{x}$, selecting a class maximizing the corresponding logit, i.e. $\hat{y} = \argmax\limits_{c \in \lbrace 1, ..., N_{classes}^{1} \rbrace} \overline{logit}_{c}(\mathbf{x})$.
In the first task, we train $a$ and $b$ to maximize a logit for the ground truth class. In all following tasks, $a$ and $b$ are \textbf{frozen}. Therefore, we just calculate logits for $N_{classes}^{:t}$ classes, where $t$ is the current task, and purely select the maximum value but we do not fine-tune these parameters, making \logit{} a computationally-efficient method.

The pseudocode of \logit{} is described in Algorithm~\ref{alg:feloc}. \logit{} requires storing a similar number of parameters like \knn{}, i.e. $N_{classes} \cdot N_{f} (N_{clusters} + N_{f}) + 2$. The only difference results from trainable $a$ and $b$. 

\section{Experiments}
\subsection{Description}
    The proposed methods, \knn{} and \logit{}\footnote{The code necessary for replicating our experiments is available at a GitHub repository: \url{https://github.com/gmum/FeNeC}.}, are considered in the Class-Incremental Learning scenario. Exemplars of previously encountered classes are not further revisited and task identity is unknown.
    

    Following FeCAM, we evaluate two feature extractors: a Vision Transformer (ViT-B/16) pre-trained on ImageNet-21K and a ResNet-18 trained on the first task samples. After training, extractor weights remain frozen. ViT produces 768-dimensional features, while ResNet-18 outputs 512-dimensional vectors. Using four datasets and two extractors, we define five experimental setups:
    
    \noindent \textbf{CIFAR-100 (ResNet / ViT)} -- It comprises 100 classes, each with 500 training and 100 test images of size $32\times32$~\cite{krizhevsky2009cifar}. For ResNet-18, the dataset is partitioned into six tasks ($T=6$), with 50 classes in the first task and 10 in each subsequent one. For ViT, CIFAR-100 is split into ten tasks ($T=10$), each containing 10 classes.

    \noindent \textbf{Tiny ImageNet (ResNet)} -- It consists of 200 classes derived from ImageNet, with images resized to $64\times64$~\cite{Le2015Tiny}. Each class contains 500 training images and 50 test images. The dataset is divided into six tasks ($T=6$), where the initial task consists of 100 classes, and the remaining five tasks each introduce 20 new classes.

    \noindent \textbf{ImageNet-Subset (ResNet)} -- This dataset is a 100-class subset of the ImageNet LSVRC dataset~\cite{Russakovsky2015ImageNet}. Most of the classes contain 1300 training images, while all have 50 test images. It is divided into six tasks following the same split strategy as CIFAR-100 on ResNet-18.
    
    
    \noindent \textbf{ImageNet-R (ViT)} -- It contains 200 classes with an unbalanced number of training and testing samples per class. The dataset is divided into ten tasks ($T=10$), each introducing 20 new classes~\cite{Wang2022dualprompt}.

    The optimal hyperparameters selected for evaluation across all experimental settings are shown in Table~\ref{tab:BestParams}. For datasets where ResNet-18 was employed as a feature extractor, sample normalization was applied when computing the Mahalanobis distance. Additionally, shrinkage of the covariance matrix was performed twice on those datasets, while once in experiments utilizing ViT. $N_{neighbors}$ is a hyperparameter of \knn{} while $N_{points}$ and $\mathrm{lr}$ (learning rate) concern \logit{}. Since Tukey's transformation may be applied only for non-negative values and ViT outputs can contain negative ones, $\lambda$ is a hyperparameter important only for datasets, where we used ResNet-18 as a feature extractor.
    
    For all experimental settings, we optimized hyperparameters using Optuna with the TPE sampler. All interval ranges selected for calculations are described in Table~\ref{tab:Searches}.

    \begin{table}[!ht]
    \centering
    \footnotesize
    \caption{Best sets of hyperparameters for each dataset
    \label{tab:BestParams}}
    
    \begin{tabular}{llllllllll}
    \toprule
    \multirow{2}{*}{$\textbf{Method}$}
     & {\textbf{Feature}}  
     & \multirow{2}{*}{\textbf{Dataset}}
     & \multirow{2}{*}{$\boldsymbol{N_{clusters}}$} 
     & {$\boldsymbol{N_{neighbors}}$} 
     & \multirow{2}{*}{$\boldsymbol{\lambda}$} 
     & \multirow{2}{*}{$\boldsymbol{\gamma_{1}}$} 
     & \multirow{2}{*}{$\boldsymbol{\gamma_{2}}$}
     & \multirow{2}{*}{\textbf{lr}}\\
     
     & {\textbf{Extractor}} & & & {or $\boldsymbol{N_{points}}$} & & & & \\ \midrule
     
    \multirow{5}{*}{\textbf{\knn{}}} 
     & \multirow{3}{*}{ResNet} & CIFAR-100 & $47$ & $1$ & $0.38$ & $0.89$ & $0.90$ & --\\[0.5ex]
    & & Tiny ImageNet & $1$ & $1$ & $0.43$ & $1.01$ & $1.32$ & --\\[0.5ex]
    & & ImageNet-Subset & $43$ & $4$ & $0.42$ & $0.92$ & $0.50$ & --\\[0.5ex]
    & \multirow{2}{*}{ViT} & CIFAR-100 & $26$ & $1$ & -- & $6.12$ & $8.10$ & --\\[0.5ex]
    & & ImageNet-R & $1$ & $1$ & -- & $9.98$ & $0.00$ & --\\ \midrule
     
    \multirow{5}{*}{\textbf{\logit{}}} 
     & \multirow{3}{*}{ResNet} & CIFAR-100 & $45$ & $2$ & $0.38$ & $1.16$ & $1.92$ & $0.00377$ \\[0.5ex]
    & & Tiny ImageNet & $24$ & $6$ & $0.45$ & $1.15$ & $1.95$ & $0.27300$ \\[0.5ex]
    & & ImageNet-Subset & $32$ & $3$ & $0.37$ & $0.90$ & $0.50$ & $0.05510$ \\[0.5ex]
    & \multirow{2}{*}{ViT} & CIFAR-100 & $50$ & $3$ & -- & $8.88$ & $12.06$ & $0.00333$ \\[0.5ex]
    & & ImageNet-R & $1$ & $1$ & -- & $10.15$ & $9.37$ & $0.14700$\\ \midrule
    \end{tabular}
\end{table}

    \setlength{\tabcolsep}{1,75mm}
    \begin{table}[!ht]
    \centering
    \small
    \caption{Domain spaces for the optimized hyperparameters. We sampled a learning rate log-uniformly within the given range.\label{tab:Searches}}

    \begin{tabular}{lllllllll}
    \toprule
     \footnotesize{\multirow{2}{*}{$\textbf{Method}$}} 
     & \footnotesize{{\textbf{Feature}}}  
     & \footnotesize{\multirow{2}{*}{\textbf{Dataset}}}
      & \scriptsize{\multirow{2}{*}{$\boldsymbol{N_{clusters}}$}} 
     & \scriptsize{$\boldsymbol{N_{neighbors}}$} 
     & \scriptsize{\multirow{2}{*}{$\boldsymbol{\lambda}$}} 
     & \scriptsize{\multirow{2}{*}{$\boldsymbol{\gamma_{1}}$}} 
     & \scriptsize{\multirow{2}{*}{$\boldsymbol{\gamma_{2}}$}}
     & \scriptsize{\multirow{2}{*}{\textbf{lr}}}\\
     
     & \footnotesize{{\textbf{Extractor}}} & & & \scriptsize{or $\boldsymbol{N_{points}}$} & & & & \\ \midrule

     
    \footnotesize{\multirow{5}{*}{\textbf{\knn{}}}} & \footnotesize{\multirow{3}{*}{ResNet}} & \footnotesize{CIFAR-100} & $\lbrace 2, 3, ..., 75 \rbrace$ & $\lbrace 1, 2, ..., 40 \rbrace$ & $[0.2, 1]$ & $[0.5, 20]$ & $[0.5, 20]$ & --\\
    
    & & \footnotesize{Tiny ImageNet} & $\lbrace 2, 3, ..., 75 \rbrace$ & $\lbrace 1, 2, ..., 40 \rbrace$ & $[0.2, 1]$ & $[0.5, 20]$ & $[0.5, 20]$ & -- \\
    
    & & \footnotesize{ImageNet-Subset} & $\lbrace 2, 3, ..., 75 \rbrace$ & $\lbrace 1, 2, ..., 40 \rbrace$ & $[0.2, 1]$ & $[0.5, 20]$ & $[0.5, 20]$ & -- \\
    
    & \footnotesize{\multirow{2}{*}{ViT}} & \footnotesize{CIFAR-100} & $\lbrace 2, 3, ..., 75 \rbrace$ & $\lbrace 1, 2, ..., 40 \rbrace$ & -- & $[0.5, 20]$ & $[0, 20]$ & -- \\
    
    & & \footnotesize{ImageNet-R} & $\lbrace 2, 3, ..., 50 \rbrace$ & $\lbrace 1, 2, ..., 40 \rbrace$ &  -- & $[0.5, 20]$ & $[0, 20]$ & --\\ \midrule


     \footnotesize{\multirow{5}{*}{\textbf{\logit{}}}} & \footnotesize{\multirow{3}{*}{ResNet}} & \footnotesize{CIFAR-100} & $\lbrace 2, 3, ..., 75 \rbrace$ & $\lbrace 1, 2, ..., 40 \rbrace$ & $[0.3, 0.6]$ & $[0.5, 20]$ & $[0.5, 20]$ & $[0.0001, 1]$ \\
    
    & & \footnotesize{Tiny ImageNet} & $\lbrace 2, 3, ..., 75 \rbrace$ & $\lbrace 1, 2, ..., 40 \rbrace$ & $[0.3, 0.6]$ & $[0.5, 20]$ & $[0.5, 20]$ & $[0.0001, 1]$ \\
    
    & & \footnotesize{ImageNet-Subset} & $\lbrace 2, 3, ..., 75 \rbrace$ & $\lbrace 1, 2, ..., 40 \rbrace$ & $[0.3, 0.6]$ & $[0.5, 20]$ & $[0.5, 20]$ & $[0.0001, 0.3]$ \\
    
    & \footnotesize{\multirow{2}{*}{ViT}} & \footnotesize{CIFAR-100} & $\lbrace 2, 3, ..., 75 \rbrace$ & $\lbrace 1, 2, ..., 40 \rbrace$ & -- & $[0.5, 20]$ & $[0.5, 20]$ & $[0.0001, 1]$ \\
    
    & & \footnotesize{ImageNet-R} & $\lbrace 2, 3, ..., 40 \rbrace$ & $\lbrace 1, 2, ..., 40 \rbrace$ & -- & $[0.5, 20]$ & $[0.5, 20]$ & $[0.0001, 1]$\\ \bottomrule

    \end{tabular}
    \end{table}

\noindent

\subsubsection{Training of \logit{}}
In the second variant of the proposed method, i.e. \logit{}, we employ two trainable parameters $a$ and $b$ which we use for calculating per-class logit values. We initialize them from a standard normal distribution, i.e. $a, b \sim \mathcal{N}(0,1)$, and then optimize using the Stochastic Gradient Descent algorithm. We achieve this, by utilizing the cross-entropy loss between the ground-truth class and predictions of the model. 

Firstly, we calculate covariance matrices and class centroids. Then, we perform training of $a$ and $b$ during the first task and after that, we freeze their values. In the case of settings with ViT, we train \logit{} maximum through 1000 epochs, and for ResNet-18 through 200 epochs. In both cases, we use a batch size of 64. We also apply early stopping, i.e. if validation loss did not decrease through 10 consecutive epochs, training is stopped.

In all cases, we use exactly the same feature extractors as FeCAM, trained as described in the corresponding manuscript.

\begin{table}[htbp]
\centering
\begin{minipage}[t]{0.54\textwidth}
    \centering
    \captionof{table}{Results of Class-Incremental Learning experiments 
      with a large initial task and five incremental tasks. For all baselines, 
      excluding FeCAM, we reported the average incremental accuracy 
      taken from~\cite{goswami2023fecam}. For FeCAM, \knn{} and \logit{}, 
      we described both the average incremental and last task accuracies, 
      averaged over three runs. We used the same three trained ResNet-18 model 
      instances as feature extractors for FeCAM, \knn{} and \logit{}. 
      $^\star$ denotes that a method converged to FeCAM, 
      i.e. best results were achieved for a single cluster and one neighbor. \label{tab:ResNet}} 
    \vspace{0.3cm}
    \begin{tabular*}{\textwidth}{@{\extracolsep{\fill}}llll@{}}
\toprule
\scriptsize{\textbf{Method}} 
& \scriptsize{\textbf{CIFAR-100}} 
& \scriptsize{\textbf{Tiny ImageNet}} 
& \scriptsize{\textbf{ImageNet-Subset}} \\
\midrule

        \multicolumn{4}{c}{Average incremental accuracy} \\ 
        \midrule 
        EWC & $24.5$ & $18.8$ & $-$ \\
        LwF-MC & $45.9$ & $29.1$ & $-$ \\
        DeeSIL & $60.0$ & $49.8$ & $67.9$ \\
        MUC & $49.4$ & $32.6$ & $-$ \\
        SDC & $56.8$ & $-$ & $-$ \\
        PASS & $63.5$ & $49.6$ & $64.4$ \\
        IL2A & $66.0$ & $47.3$ & $-$ \\
        SSRE & $65.9$ & $50.4$ & $-$ \\
        FeTrIL & $67.6$ & $55.4$ & $73.1$ \\ 
        \midrule
        FeCAM & $71.06$\tiny{$\pm 0.5$} & $60.02$\tiny{$\pm 0.6$} & $78.49$\tiny{$\pm 0.5$} \\
        \knn{} & $71.90$\tiny{$\pm 0.8$} & $60.19$\tiny{$\pm 0.4^\star$} 
               & $\mathbf{79.59}$\tiny{$\mathbf{\pm 0.5}$}  \\
        \logit{} & $\mathbf{71.94}$\tiny{$\mathbf{\pm 0.8}$} 
                  & $\mathbf{60.23}$\tiny{$\mathbf{\pm 0.3}$} 
                  & $79.56$\tiny{$\pm 0.4$} \\ 
        \midrule
        \multicolumn{4}{c}{Last task accuracy} \\ 
        \midrule
        FeCAM & $61.99$\tiny{$\pm 0.3$} & $52.63$\tiny{$\pm 0.4$} & $70.97$\tiny{$\pm 0.8$} \\
        \knn{} & $63.30$\tiny{$\pm 0.4$} 
                & $\mathbf{52.93}$\tiny{$\mathbf{\pm 0.3}$}$^\star$ 
                & $72.49$\tiny{$\pm 0.9$} \\
        \logit{} & $\mathbf{63.60}$\tiny{$\mathbf{\pm 0.5}$} 
                  & $52.87$\tiny{$\pm 0.3$} 
                  & $\mathbf{72.53}$\tiny{$\mathbf{\pm 0.7}$} \\ 
        \bottomrule
    \end{tabular*}
\end{minipage}
\hfill
\begin{minipage}[t]{0.43\textwidth}
    \centering
    \captionof{table}{Results of Class-Incremental Learning experiments 
      with ten equally-sized tasks. For baselines, excluding FeCAM, we reported 
      the last task accuracy taken from~\cite{goswami2023fecam}. For FeCAM, 
      \knn{} and \logit{}, we described both the last task and average 
      incremental accuracies, averaged over three runs. We used the ViT 
      architecture pre-trained on ImageNet-21K model as feature extractors for 
      FeCAM, \knn{} and \logit{}. $^\star$ denotes that a method converged to 
      FeCAM, i.e. best results were achieved for a single cluster and one 
      neighboring point. \label{tab:ViT}} 
      
    \vspace{0.3cm}
    \begin{tabular*}{\linewidth}{@{\extracolsep{\fill}} l l l @{}}
        \toprule
        \textbf{\scriptsize{Method}} 
          & \textbf{\scriptsize{CIFAR-100}} 
          & \textbf{\scriptsize{ImageNet-R}} \\ 
        \midrule
        \multicolumn{3}{c}{Last task accuracy} \\ 
        \midrule
        FT-frozen & $17.7$ & $39.5$ \\
        FT & $33.6$ & $28.9$ \\
        EWC & $47.0$ & $35.0$ \\
        LwF & $60.7$ & $38.5$ \\
        L2P & $83.8$ & $61.6$ \\
        NCM & $83.7$ & $55.7$ \\ 
        \midrule
        FeCAM & $85.67$\tiny{$\pm 0.0$} & $64.04$\tiny{$\pm 0.3$}  \\
        \knn{} & $87.17$\tiny{$\pm 0.0$} 
                & $\mathbf{64.07}$\tiny{$\mathbf{\pm 0.3}$}$^{\star}$ \\
        \logit{} & $\mathbf{87.21}$\tiny{$\mathbf{\pm 0.1}$} 
                  & $\mathbf{64.07}$\tiny{$\mathbf{\pm 0.3}$}$^{\star}$ \\ 
        \midrule
        \multicolumn{3}{c}{Average incremental accuracy} \\ 
        \midrule
        FeCAM & $89.61$\tiny{$\pm 0.4$} & $70.35$\tiny{$\pm 0.5$} \\
        \knn{} & $\mathbf{91.12}$\tiny{$\mathbf{\pm 0.3}$} & $70.36$\tiny{$\pm 0.5$} \\
        \logit{} & $91.11$\tiny{$\pm 0.3$} & $\mathbf{70.38}$\tiny{$\mathbf{\pm 0.5}$} \\ 
        \bottomrule
    \end{tabular*}
\end{minipage}
\end{table}


\subsection{Results}

The results presented in Tables \ref{tab:ResNet}--\ref{tab:ViT} demonstrate that both \knn{} and \logit{} consistently achieve superior performance across several experimental configurations, both in average incremental accuracy and last-task accuracy. On Tiny ImageNet and ImageNet-R datasets, our methods converge to FeCAM, yielding only marginal improvements.

According to Table \ref{tab:ResNet}, \knn{} and \logit{} with a frozen ResNet-18 backbone exhibit significant performance gains over other methods, with only FeCAM remaining competitive. The most pronounced improvements occur on ImageNet-Subset, where both \knn{} and \logit{} outperform FeCAM by over 1\%. Since ImageNet-Subset is the largest dataset tested, these results support the conclusion that our methods effectively handle large-scale continual learning tasks. On CIFAR-100, we also ob-

   serve a visible improvement over FeCAM, whereas on Tiny ImageNet, \knn{} converges to FeCAM, meaning that the slight improvement stems solely from adjustments to $\gamma_{1}$, $\gamma_{2}$, and $\lambda$.

   A similar trend is observed in Table \ref{tab:ViT}, where \knn{} and \logit{} outperform all baselines on CIFAR-100 (ViT), reaching 87.17\% and 87.21\%, respectively, compared to 85.67\% for FeCAM -- a notable gain of over 1.5\%. However, on ImageNet-R, our methods again converge to FeCAM. Notably, ImageNet-R, despite having the highest number of classes among the tested datasets, contains relatively few samples per class. Many classes have fewer than 100 examples, whereas, on ImageNet-Subset, nearly all classes contain 1,300 training samples. 
It is likely that for more complex datasets like ImageNet-Subset, detecting multiple centroids better captures the underlying data structure, improving adaptation to class diversity.

\begin{figure*}[!ht]
    \centering
        \begin{subfigure}{0.33\linewidth}
            \includegraphics[trim={0,0cm, 0,0cm, 0,0cm, 0,0cm},clip,width=\linewidth]{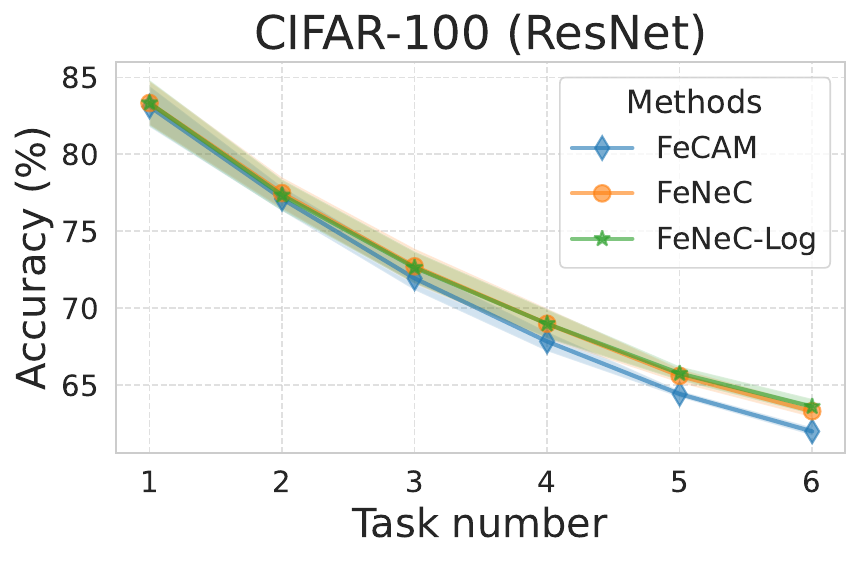}
        \end{subfigure}%
        \begin{subfigure}{0.33\linewidth}
            \includegraphics[trim={0,0cm, 0,0cm, 0,0cm, 0,0cm},clip,width=\linewidth]{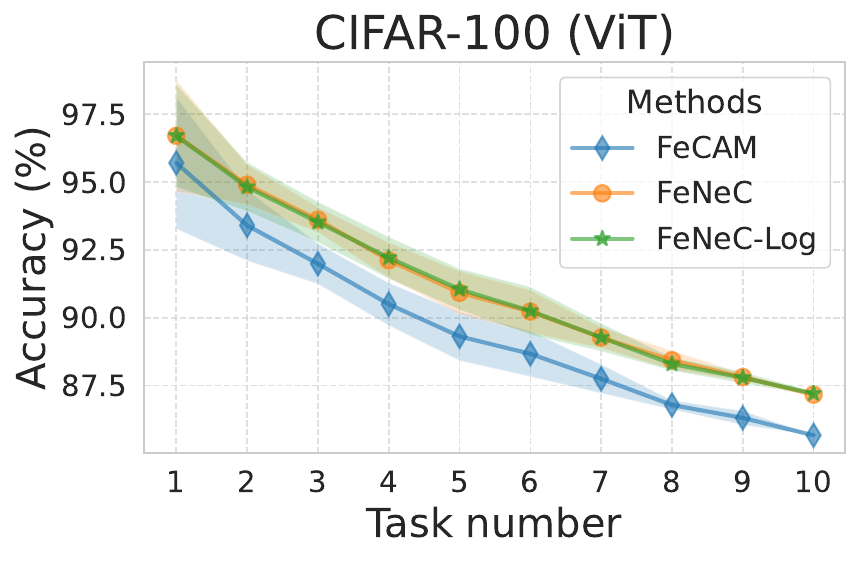}
        \end{subfigure}
        \begin{subfigure}{0.33\linewidth}
            \includegraphics[trim={0,0cm, 0,0cm, 0,0cm, 0,0cm},clip,width=\linewidth]{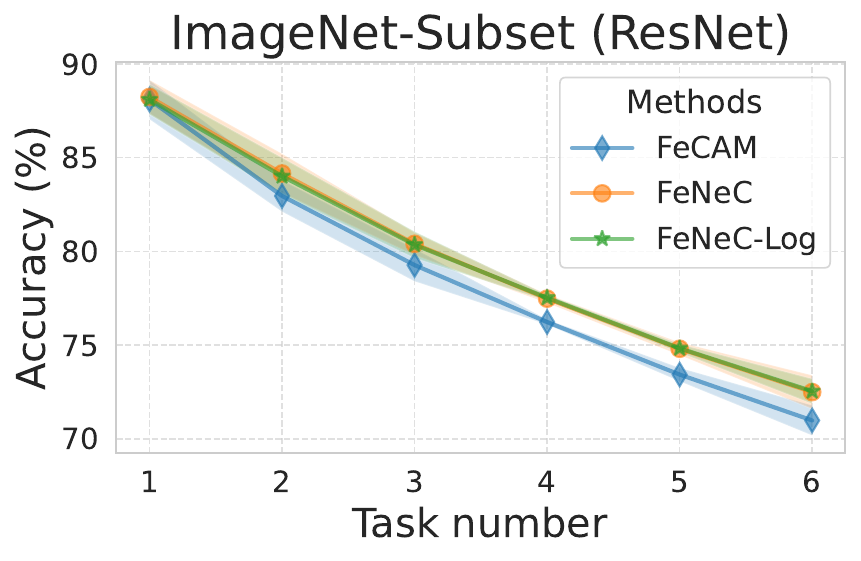}
        \end{subfigure}
    \caption{The comparison of FeCAM, \knn{} and \logit{} across incremental tasks on different datasets. The results were averaged over three different class orders (and random initializations for ResNet) for all methods.
    \label{results:task_accuracy}}
\end{figure*}

Figure \ref{results:task_accuracy} illustrates the mean accuracy (with 95\% confidence intervals) of FeCAM, \knn{}, and \logit{} across backbone models trained in tasks with three different class orders and -- in the case of ResNet-18 -- network initializations. The results indicate that \logit{} and \knn{} consistently maintain higher accuracy across tasks, outperforming FeCAM in three cases across different random initializations. Notably, while initial task accuracy is often similar, FeCAM exhibits a more pronounced decline in later tasks. As expected, accuracy decreases as more tasks are introduced. However, \knn{} and \logit{} consistently retain higher accuracy compared to FeCAM, suggesting improved knowledge retention.

This effect is particularly noticeable on CIFAR-100 and ImageNet-Subset, where FeCAM exhibits a sharper accuracy decline over tasks, while \knn{} and \logit{} demonstrate greater stability. For instance, on ImageNet-Subset, the difference between first-task and last-task accuracy is, on average, 15.6\% for \logit{}, compared to 17.1\% for FeCAM. 

Across all tested configurations, \logit{} and \knn{} achieve comparable results, with \logit{} slightly outperforming \knn{} in most cases. The primary advantage of \knn{} lies in its computational efficiency and straightforward interpretability. However, \logit{} uniquely produces probability distributions over classes, allowing for further analysis of logit value distributions. This property may sometimes make \logit{} preferable over \knn{}.

\subsubsection{Comparison with FeCAM}
Our experiments were conducted across three random seeds, confirming the consistency of our methods in achieving superior results and outperforming existing approaches. Notably, \logit{} and \knn{} consistently surpass FeCAM, particularly in mitigating accuracy degradation over successive tasks. These improvements are consistent across different backbone architectures (ResNet, ViT) and datasets, underscoring the robustness of our approach in continual learning scenarios.

Additionally, we evaluated \knn{} with $N_{clusters} = N_{neighbors} = 1$, while keeping all other hyperparameters identical to those in the best-performing configurations. This setup resulted in an accuracy increase compared to the reproduction of FeCAM's reported hyperparameters. However, the best-performing \knn{} variant still achieves superior results, with the final task accuracy showing an average improvement of 0.4\% for CIFAR-100 (pretrained on ViT), 0.5\% for ImageNet-Subset, and up to 0.8\% for CIFAR-100 with ResNet.


\subsection{Hyperparameter evaluation}
\begin{figure*}[!ht]
    \centering
    \begin{subfigure}[c]{\textwidth}
        \centering
        \includegraphics[trim={0,0cm, 0,0cm, 0,0cm, 0,0cm},clip,width=0.33\linewidth]{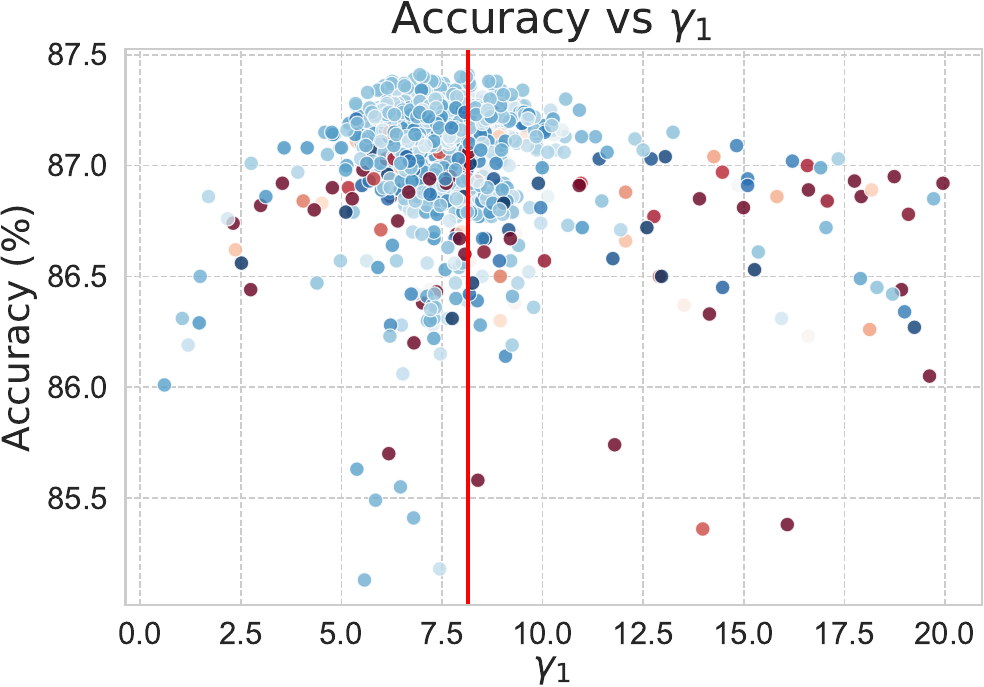}%
        \hspace{0.05cm}%
        \includegraphics[trim={0,0cm, 0,0cm, 0,0cm, 0,0cm},clip,width=0.33\linewidth]{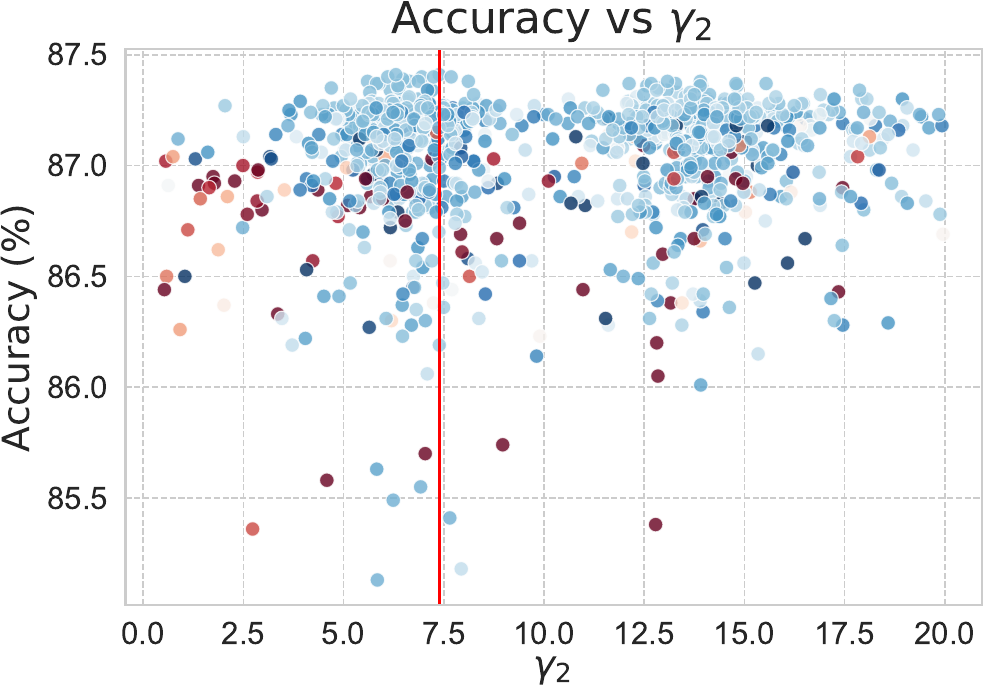}%
        \hspace{0.05cm}%
        \includegraphics[trim={0,0cm, 0,0cm, 0,0cm, 0,0cm},clip,width=0.33\linewidth]{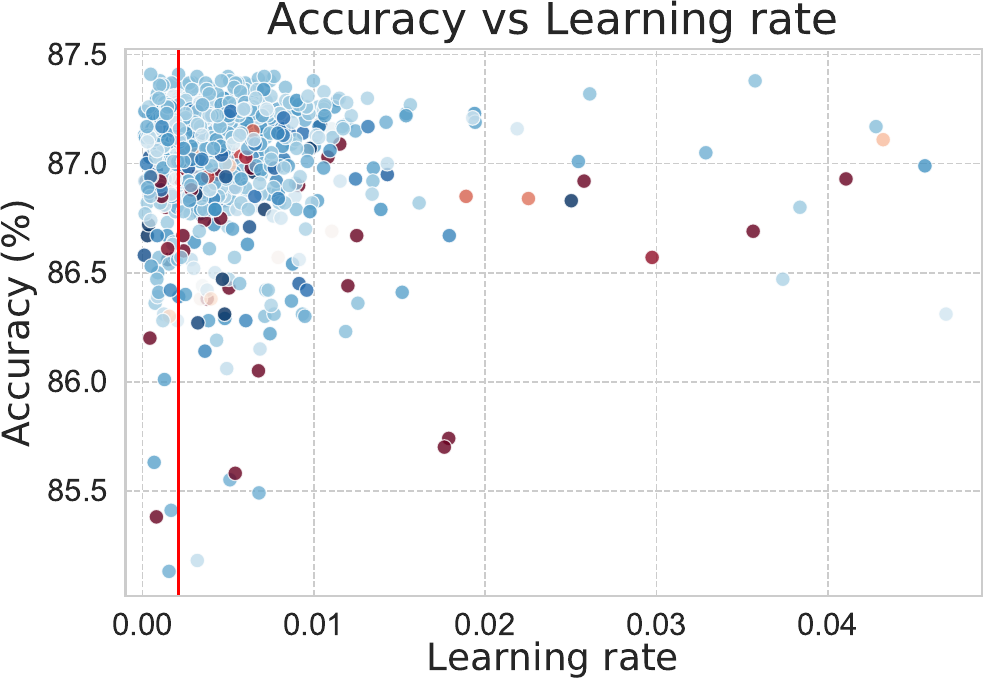}
        \includegraphics[trim={0,0cm, 0,0cm, 0,0cm, 0,0cm},clip,width=0.4\linewidth]{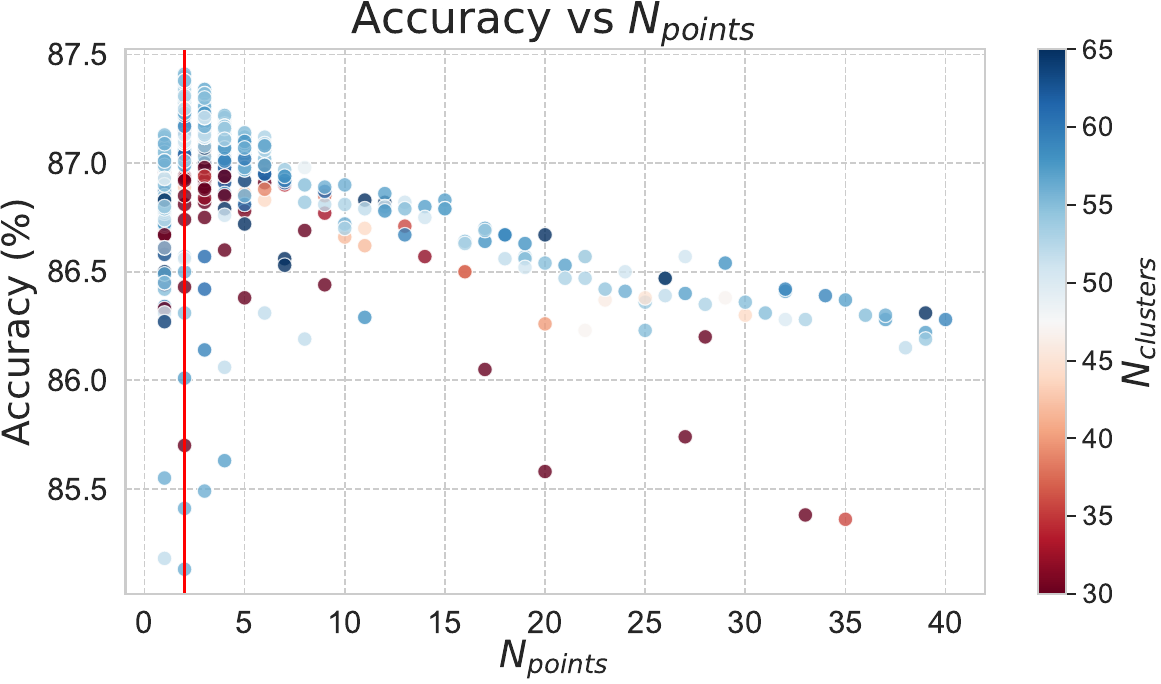}%
        \hspace{0,15cm}%
        \includegraphics[trim={0,0cm, 0,0cm, 0,0cm, 0,0cm},clip,width=0.4\linewidth]{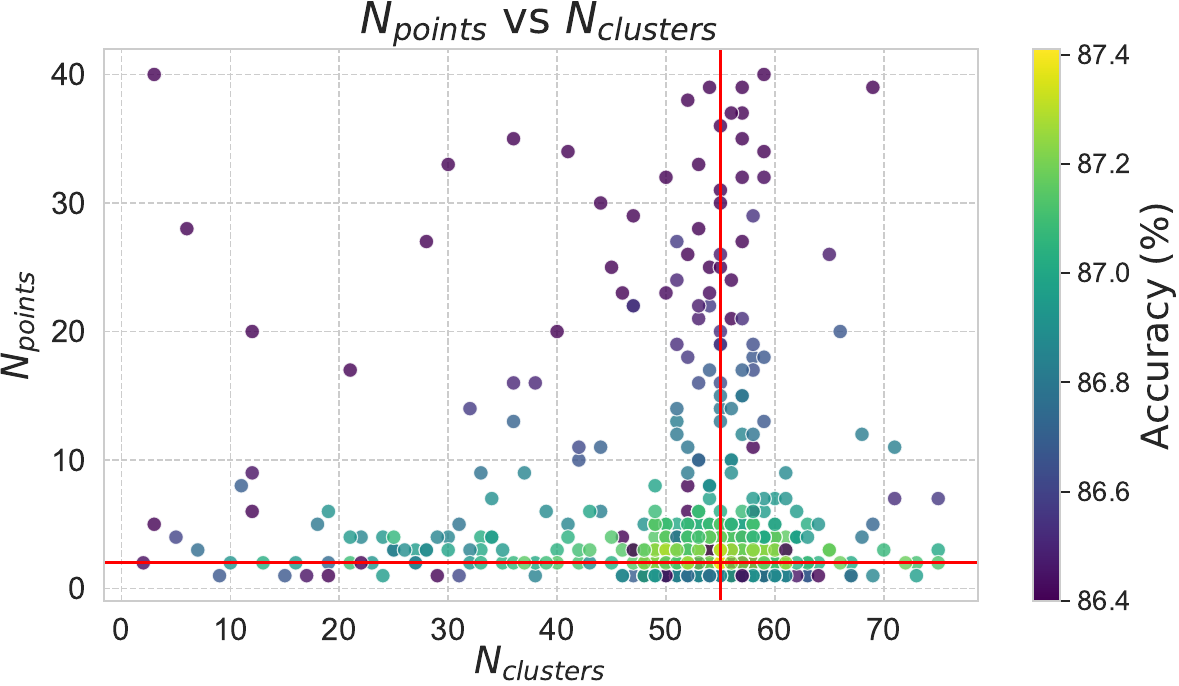}
        \caption{CIFAR-100}\label{ablation:ViT_CIFAR}
    \end{subfigure}
    \begin{subfigure}[c]{\textwidth}
        \centering
        \includegraphics[trim={0,0cm, 0,0cm, 0,0cm, 0,0cm},clip,width=0.33\linewidth]{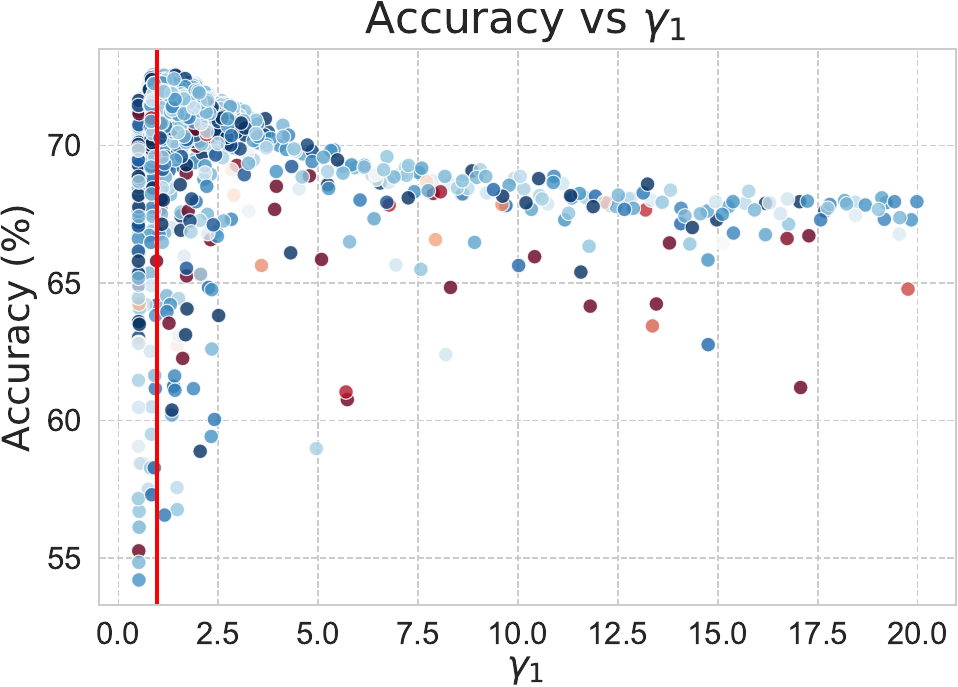}%
        \hspace{0.05cm}%
        \includegraphics[trim={0,0cm, 0,0cm, 0,0cm, 0,0cm},clip,width=0.33\linewidth]{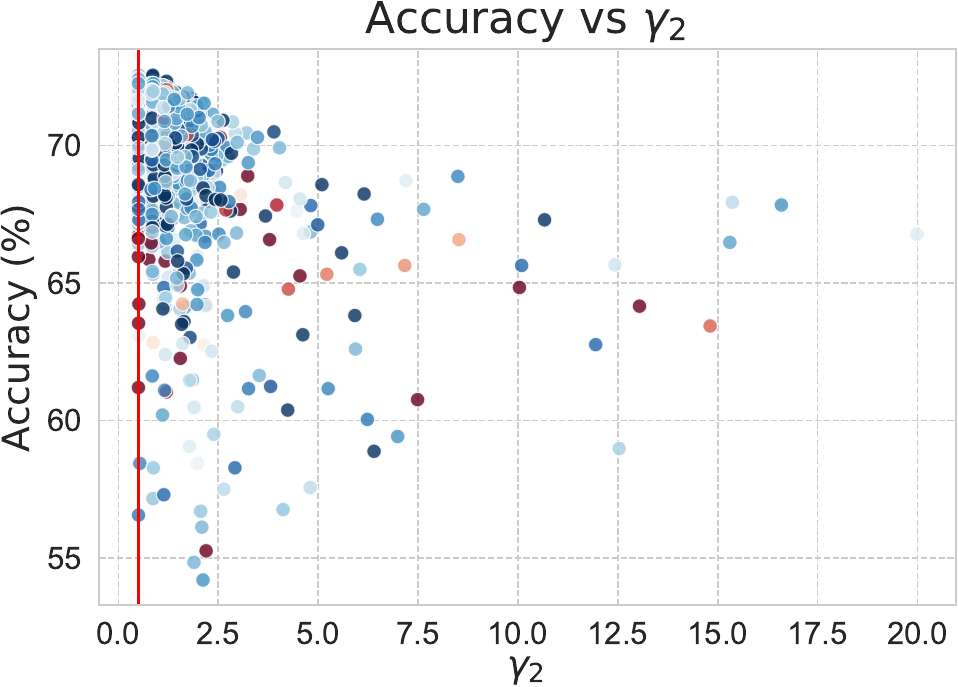}%
        \hspace{0.05cm}%
        \includegraphics[trim={0,0cm, 0,0cm, 0,0cm, 0,0cm},clip,width=0.33\linewidth]{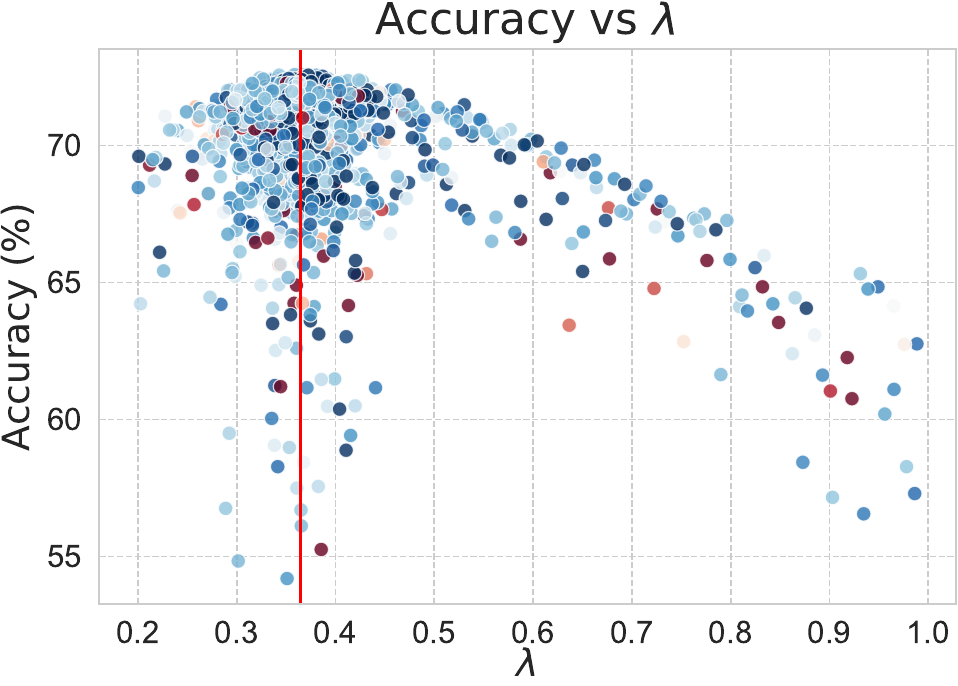}
        \includegraphics[trim={0,0cm, 0,0cm, 0,0cm, 0,0cm},clip,width=0.4\linewidth]{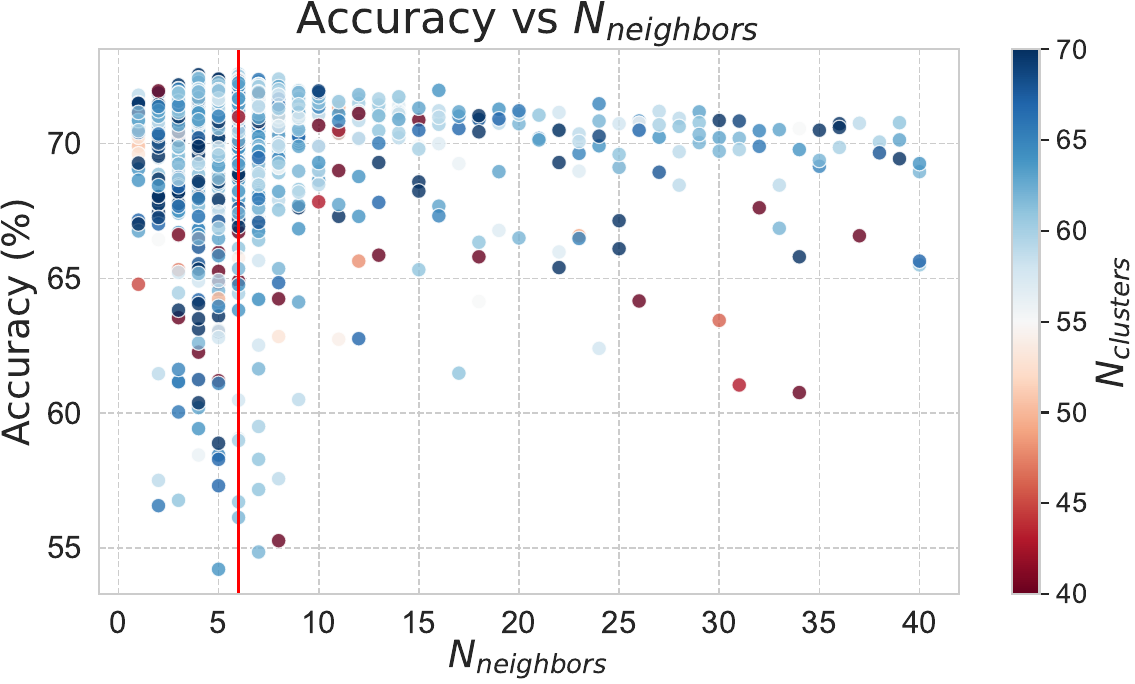}%
        \hspace{0,15cm}%
        \includegraphics[trim={0,0cm, 0,0cm, 0,0cm, 0,0cm},clip,width=0.4\linewidth]{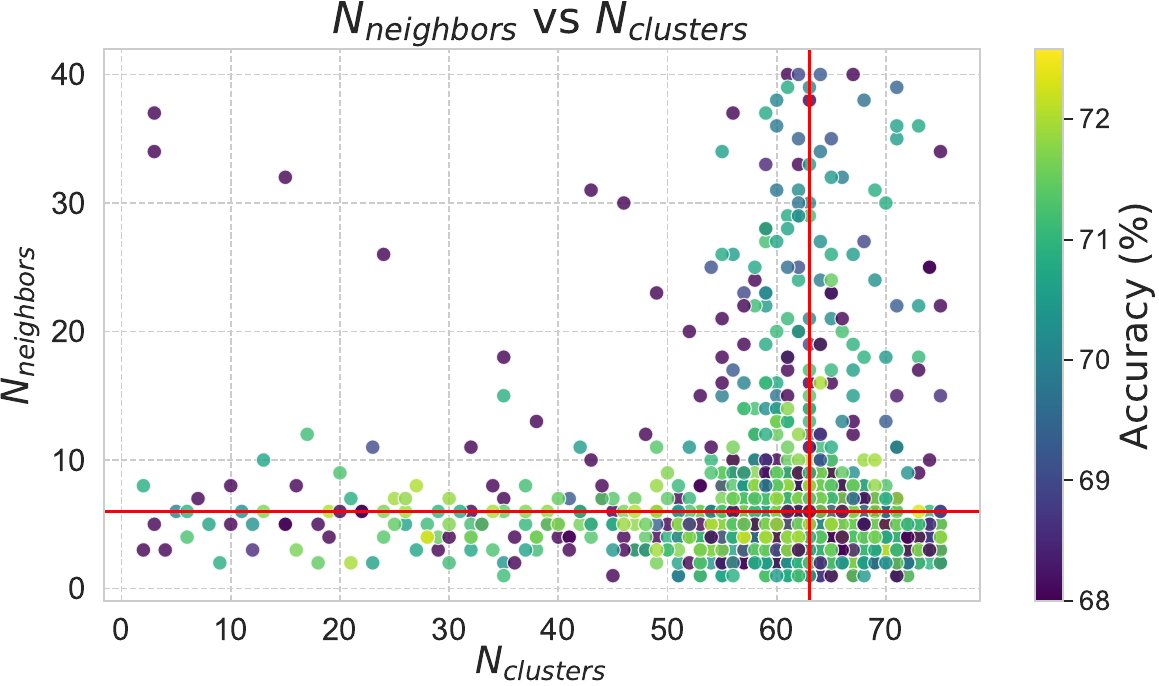}
    \caption{ImageNet-Subset}\label{ablation:ResNet_ImageNetSubset}
    \end{subfigure}
    \caption{The last task accuracy for different hyperparameter values of \knn{} with ResNet-18 architecture and the ImageNet-Subset dataset. The best hyperparameters are marked with red lines. In the graphs comparing accuracy with hyperparameters, the color indicates the number of clusters, while in the graph comparing $N_{points}$ or $N_{neighbors}$ with $N_{clusters}$, the color indicates accuracy.\label{ablation:general_hyperparams}}
\end{figure*}

Figure \ref{ablation:general_hyperparams} shows hyperparameter optimization results using Optuna~\cite{optuna2019} with the TPE sampler. This method favors promising regions, leading to clustered successful configurations rather than a uniform search space distribution.

For each dataset, the number of selected clusters, $N_{clusters}$, does not exceed 10\% of the total number of samples per class. This balance allows our methods to capture a rich representation of the dataset while preventing excessive retention of centroids from previous tasks.


Figure \ref{ablation:general_hyperparams} illustrates that, for ImageNet-Subset and CIFAR-100 (ViT), increasing the number of clusters to 45--70 leads to better performance. However, this trend does not hold for ImageNet-R, suggesting that its intrinsic characteristics favor a smaller number of clusters. For ImageNet-Subset and CIFAR-100 (ViT), using fewer than 10 neighbors ensures the highest accuracy. In the case of the representations extracted from ViT, $\gamma_1$ converges to high values and $\gamma_2$ remains scattered throughout the studied domain, indicating its low relevance for this data representation (Figure \ref{ablation:ViT_CIFAR}). In contrast, for the representations obtained from ResNet-18 (Figure~\ref{ablation:ResNet_ImageNetSubset}), both hyperparameters are significant and converge to low values.

\begin{table}
    \centering
    \caption{Average incremental accuracy for Class-Incremental Learning experiments for \knn{} and \logit{}, for two different distance metrics: Mahalanobis (default in our methods), denoted as (M) and Euclidean distance (E) for the comparison purpose. Results are averaged over three runs. $^\star$ denotes that a method converged to FeCAM, i.e. best results were achieved for a single cluster and one neighboring point.\label{ablation:metric}}

    \begin{tabular*}{\textwidth}{@{\extracolsep{\fill}} llllll }
        \toprule
         \multirow{2}{*}{\textbf{Method}} & \textbf{CIFAR-100} & \textbf{CIFAR-100} & \textbf{Tiny} & \textbf{ImageNet-} & \multirow{2}{*}{\textbf{ImageNet-R}} \\
         & \textbf{(ViT)} & \textbf{(ResNet)} & \textbf{ImageNet} & \textbf{Subset} & \\
         \midrule
         \knn{} (E)   & $88.74 \pm 0.3$   & $67.99 \pm 0.9$   & $54.25 \pm 0.4^\star$ & $76.88 \pm 0.2$ & $63.28 \pm 1.2^\star$ \\
         \knn{} (M)   & $91.12 \pm 0.3$   & $71.90 \pm 0.8$   & $60.19 \pm 0.4^\star$ & $79.59 \pm 0.5$ & $70.36 \pm 0.5^\star$ \\
         \midrule
         \logit{} (E) & $90.31 \pm 0.1$   & $68.38 \pm 1.0$   & $56.08 \pm 0.8$      & $76.73 \pm 0.0$ & $63.28 \pm 1.2^\star$ \\
         \logit{} (M) & $91.11 \pm 0.3$   & $71.94 \pm 0.8$   & $60.23 \pm 0.3$      & $79.56 \pm 0.4$ & $70.38 \pm 0.5^\star$ \\
        \bottomrule     
    \end{tabular*}
\end{table}

Furthermore, we experimented with training \logit{} on all tasks simultaneously, incorporating regularization to prevent large parameter shifts between tasks. However, the results remained similar while the training time increased significantly. This suggests that our current approach effectively balances performance and computational efficiency.

\subsection{Distance metric}

As shown in Table \ref{ablation:metric}, using the Euclidean distance results in visibly worse performance compared to the Mahalanobis distance across all datasets. The performance gap is particularly pronounced on ImageNet-R, where the Mahalanobis distance provides an improvement of up to 7\%, highlighting its advantages. Despite this, our approaches still achieve competitive performance, when comparing the Euclidean results with average incremental accuracy of methods presented in Table \ref{tab:ResNet}. Aside from FeCAM, which also leverages the Mahalanobis distance, \knn{} and \logit{} with the Euclidean distance outperform all other baselines in 3 out of 3 settings. The key advantage of the Euclidean distance is its lower computational cost, as it eliminates the need to compute and store covariance matrices for each class, making it a more efficient alternative in resource-constrained environments.

\subsection{Runtime}
We present runtime evaluation results for \knn{} and \logit{} in Table~\ref{tab:Runtime}. The reported values represent the average runtime with standard deviations only for the training of \knn{} and \logit{}, without the learning of the feature extractors. The experiments were performed on the NVIDIA GeForce RTX 2080 Ti graphic card.

\begin{table}[!ht]
    \centering
    \caption{\knn{} and \logit{} runtime on each dataset (without learning of the feature extractor}
    \label{tab:Runtime}
    
    \begin{tabular}{lll}
    \toprule
    \textbf{Dataset} & \textbf{FeNeC} & \textbf{FeNeC-Log} \\ \midrule
    
    \textbf{CIFAR-100 (ResNet)} & 
    $0~\mathrm{min}~24~s \pm 0.1~s$ & 
    $2~\mathrm{min}~29~s \pm 8.8~s$ \\ 
    
    \textbf{Tiny ImageNet} & 
    $0~\mathrm{min}~9~s \pm 0.1~s$ & 
    $1~\mathrm{min}~04~s \pm 10.7~s$ \\ 
    
    \textbf{ImageNet-Subset} & 
    $0~\mathrm{min}~27~s \pm 1.6~s$ & 
    $4~\mathrm{min}~25~s \pm 15.5~s$ \\ 
    
    \textbf{CIFAR-100 (ViT)} & 
    $0~\mathrm{min}~30~s \pm 0.2~s$ & 
    $2~\mathrm{min}~50~s \pm 1.9~s$ \\ 
    
    \textbf{ImageNet-R} & 
    $0~\mathrm{min}~14~s \pm 0.1~s$ & 
    $1~\mathrm{min}~11~s \pm 11.6~s$ \\ 
    
    \bottomrule
    \end{tabular}
\end{table}

\section{Conclusion and Future Works}
In this work, we propose \knn{} which is an extension of a single prototype approach per class into multiple cluster representation. This enables better adaptation to diverse data. Then, we classify samples using the nearest neighbor technique. Moreover, \logit{} calculates a log-likelihood function with two trainable parameters, shared among classes. Our solution generalizes FeCAM, i.e. with a single cluster and single neighboring point, \knn{} reduces to FeCAM. We evaluated two variants of \knn{} in several benchmarks, in most cases improving FeCAM results. In some situations, creating a single cluster remains the most efficient setting. In future works, one may examine different cluster selection techniques or modify logits creating more sophisticated equations and per-class sets of parameters.

\subsection*{Acknowledgments}
The work of Kamil Książek and Jacek Tabor, as well as the infrastructure for experiments, were funded from the flagship project entitled ''Artificial Intelligence Computing Center Core Facility'' from the DigiWorld Priority Research Area within the Excellence Initiative – Research University program at Jagiellonian University in Krakow.

\subsection*{Disclosure of Interests.}
The authors have no competing interests to declare that are
relevant to the content of this article.

%
%
%
\bibliographystyle{splncs04}
\bibliography{bibliography}

\begin{thebibliography}{10}
\providecommand{\url}[1]{\texttt{#1}}
\providecommand{\urlprefix}{URL }
\providecommand{\doi}[1]{https://doi.org/#1}

\bibitem{abraham2005MR}
Abraham, W.C., Robins, A.: Memory retention – the synaptic stability versus
  plasticity dilemma. Trends in Neurosciences  \textbf{28}(2),  73--78 (2005).
  \doi{10.1016/j.tins.2004.12.003}

\bibitem{ahn2019UCL}
Ahn, H., Lee, D., Cha, S., Moon, T.: Uncertainty-based continual learning with
  adaptive regularization (05 2019). \doi{10.48550/arXiv.1905.11614}

\bibitem{optuna2019}
Akiba, T., Sano, S., Yanase, T., Ohta, T., Koyama, M.: Optuna: A
  next-generation hyperparameter optimization framework. In: Proceedings of the
  25th {ACM} {SIGKDD} International Conference on Knowledge Discovery and Data
  Mining (2019)

\bibitem{aljundi2018MAS}
Aljundi, R., Babiloni, F., Elhoseiny, M., Rohrbach, M., Tuytelaars, T.: Memory
  aware synapses: Learning what (not) to forget. In: Proceedings of the
  European conference on computer vision (ECCV). pp. 139--154 (2018)

\bibitem{ayub2021eec}
Ayub, A., Wagner, A.R.: Eec: Learning to encode and regenerate images for
  continual learning. arXiv preprint arXiv:2101.04904  (2021)

\bibitem{chaudhry2018agem}
Chaudhry, A., Ranzato, M., Rohrbach, M., Elhoseiny, M.: Efficient lifelong
  learning with a-gem. In: International Conference on Learning Representations
  (2019)

\bibitem{cunningham2022KNN}
Cunningham, P., Delany, S.J.: k-nearest neighbour classifiers - a tutorial. ACM
  Comput. Surv.  \textbf{54}(6) (Jul 2021). \doi{10.1145/3459665}

\bibitem{delange2019CL}
De~Lange, M., Aljundi, R., Masana, M., Parisot, S., Jia, X., Leonardis, A.,
  Slabaugh, G., Tuytelaars, T.: Continual learning: A comparative study on how
  to defy forgetting in classification tasks. arXiv preprint 1909.08383  (2019)

\bibitem{frankle2018lth}
Frankle, J., Carbin, M.: The lottery ticket hypothesis: Finding sparse,
  trainable neural networks. In: International Conference on Learning
  Representations (2019), \url{https://openreview.net/forum?id=rJl-b3RcF7}

\bibitem{goswami2023fecam}
Goswami, D., Liu, Y., Twardowski, B., van~de Weijer, J.: Fecam: Exploiting the
  heterogeneity of class distributions in exemplar-free continual learning. In:
  Advances in Neural Information Processing Systems (NeurIPS) (2023)

\bibitem{hemati2023partial}
Hemati, H., Lomonaco, V., Bacciu, D., Borth, D.: Partial hypernetworks for
  continual learning. In: Conference on Lifelong Learning Agents. pp. 318--336.
  PMLR (2023)

\bibitem{henning2021posterior}
Henning, C., Cervera, M., D'Angelo, F., Von~Oswald, J., Traber, R., Ehret, B.,
  Kobayashi, S., Grewe, B.F., Sacramento, J.: Posterior meta-replay for
  continual learning. Advances in neural information processing systems
  \textbf{34},  14135--14149 (2021)

\bibitem{hsu2018RCLS}
Hsu, Liu, Ramasamy, Kira: Re-evaluating continual learning scenarios: A
  categorization and case for strong baselines. In: NeurIPS Continual Learning
  Workshop (2018)

\bibitem{janson2023simplebaseline}
Janson, P., Zhang, W., Aljundi, R., Elhoseiny, M.: A simple baseline that
  questions the use of pretrained-models in continual learning. In: Workshop on
  Distribution Shifts, , NeurIPS 2022 (2023)

\bibitem{kang2022WSN}
Kang, H., Mina, R.J.L., Madjid, S.R.H., Yoon, J., Hasegawa-Johnson, M., Hwang,
  S.J., Yoo, C.D.: Forget-free continual learning with winning subnetworks. In:
  Chaudhuri, K., Jegelka, S., Song, L., Szepesvari, C., Niu, G., Sabato, S.
  (eds.) Proceedings of the 39th International Conference on Machine Learning.
  Proceedings of Machine Learning Research, vol.~162, pp. 10734--10750. PMLR
  (17--23 Jul 2022), \url{https://proceedings.mlr.press/v162/kang22b.html}

\bibitem{kirkpatric2017OCF}
Kirkpatrick, J., Pascanu, R., Rabinowitz, N., Veness, J., Desjardins, G., Rusu,
  A.A., Milan, K., Quan, J., Ramalho, T., Grabska-Barwinska, A., Hassabis, D.,
  Clopath, C., Kumaran, D., Hadsell, R.: Overcoming catastrophic forgetting in
  neural networks. Proceedings of the National Academy of Sciences
  \textbf{114}(13),  3521--3526 (2017). \doi{10.1073/pnas.1611835114}

\bibitem{kong2024hybrid}
Kong, J., Shi, J., Gao, A., Hu, S., Zhou, T., Shao, H.: Hybrid memory replay:
  Blending real and distilled data for class incremental learning. arXiv
  preprint arXiv:2410.15372  (2024)

\bibitem{krizhevsky2009cifar}
Krizhevsky, A.: Learning multiple layers of features from tiny images.
  University of Toronto  (2009)

\bibitem{książek2023hypermask}
Książek, K., Spurek, P.: Hypermask: Adaptive hypernetwork-based masks for
  continual learning (2023)

\bibitem{Le2015Tiny}
Le, Y., Yang, X.S.: Tiny imagenet visual recognition challenge (2015)

\bibitem{lin2022trgp}
Lin, S., Yang, L., Fan, D., Zhang, J.: {TRGP}: Trust region gradient projection
  for continual learning. In: International Conference on Learning
  Representations (2022), \url{https://openreview.net/forum?id=iEvAf8i6JjO}

\bibitem{liu2022few}
Liu, H., Gu, L., Chi, Z., Wang, Y., Yu, Y., Chen, J., Tang, J.: Few-shot
  class-incremental learning via entropy-regularized data-free replay. In:
  European Conference on Computer Vision. pp. 146--162. Springer (2022)

\bibitem{lopez2017GEM}
Lopez-Paz, D., Ranzato, M.: Gradient episodic memory for continual learning.
  In: Proceedings of the 31st International Conference on Neural Information
  Processing Systems. p. 6470–6479. NIPS'17, Curran Associates Inc., Red
  Hook, NY, USA (2017)

\bibitem{ma2023progressive}
Ma, C., Ji, Z., Huang, Z., Shen, Y., Gao, M., Xu, J.: Progressive voronoi
  diagram subdivision enables accurate data-free class-incremental learning.
  In: The Eleventh International Conference on Learning Representations (2023),
  \url{https://openreview.net/forum?id=zJXg_Wmob03}

\bibitem{mallya2018packnet}
Mallya, A., Lazebnik, S.: Packnet: Adding multiple tasks to a single network by
  iterative pruning. In: Proceedings of the IEEE conference on Computer Vision
  and Pattern Recognition. pp. 7765--7773 (2018)

\bibitem{masana2023CIL}
Masana, M., Liu, X., Twardowski, B., Menta, M., Bagdanov, A.D., van~de Weijer,
  J.: Class-incremental learning: Survey and performance evaluation on image
  classification. IEEE Transactions on Pattern Analysis and Machine
  Intelligence  \textbf{45}(5),  5513--5533 (2023).
  \doi{10.1109/TPAMI.2022.3213473}

\bibitem{mccloskey1989CIiCN}
McCloskey, M., Cohen, N.J.: Catastrophic interference in connectionist
  networks: The sequential learning problem. In: Bower, G.H. (ed.) Psychology
  of Learning and Motivation, Psychology of Learning and Motivation, vol.~24,
  pp. 109--165. Academic Press (1989). \doi{10.1016/S0079-7421(08)60536-8}

\bibitem{oswald2019hypercl}
von Oswald, J., Henning, C., Grewe, B.F., Sacramento, J.: Continual learning
  with hypernetworks. In: International Conference on Learning Representations
  (2020), \url{https://arxiv.org/abs/1906.00695}

\bibitem{petit2023fetril}
Petit, G., Popescu, A., Schindler, H., Picard, D., Delezoide, B.: Fetril:
  Feature translation for exemplar-free class-incremental learning. In:
  Proceedings of the IEEE/CVF winter conference on applications of computer
  vision. pp. 3911--3920 (2023)

\bibitem{rios2018closed}
Rios, A., Itti, L.: Closed-loop memory gan for continual learning. arXiv
  preprint arXiv:1811.01146  (2018)

\bibitem{Russakovsky2015ImageNet}
Russakovsky, O., Deng, J., Su, H., Krause, J., Satheesh, S., Ma, S., Huang, Z.,
  Karpathy, A., Khosla, A., Bernstein, M., Berg, A.C., Fei-Fei, L.: Imagenet
  large scale visual recognition challenge. International Journal of Computer
  Vision  \textbf{115}(3),  211–252 (Dec 2015).
  \doi{10.1007/s11263-015-0816-y}

\bibitem{rypesc2025task}
Rype{\'s}{\'c}, G., Cygert, S., Trzcinski, T., Twardowski, B.: Task-recency
  bias strikes back: Adapting covariances in exemplar-free class incremental
  learning. Advances in Neural Information Processing Systems  \textbf{37},
  63268--63289 (2025)

\bibitem{schmidhuber2015DL}
Schmidhuber, J.: Deep learning in neural networks: An overview. Neural Networks
   \textbf{61},  85--117 (2015).
  \doi{https://doi.org/10.1016/j.neunet.2014.09.003}

\bibitem{shin2017continual}
Shin, H., Lee, J.K., Kim, J., Kim, J.: Continual learning with deep generative
  replay. Advances in neural information processing systems  \textbf{30} (2017)

\bibitem{tang2021layerwise}
Tang, S., Chen, D., Zhu, J., Yu, S., Ouyang, W.: Layerwise optimization by
  gradient decomposition for continual learning. In: Proceedings of the
  IEEE/CVF conference on Computer Vision and Pattern Recognition. pp.
  9634--9643 (2021)

\bibitem{Vandeven2022scenarios}
van~de Ven, G.M., Tuytelaars, T., Tolias, A.S.: Three types of incremental
  learning. Nature Machine Intelligence  \textbf{4}(12),  1185--1197 (2022).
  \doi{10.1038/s42256-022-00568-3}

\bibitem{verwimp2024CLApplications}
Verwimp, E., Aljundi, R., Ben-David, S., Bethge, M., Cossu, A., Gepperth, A.,
  Hayes, T.L., H{\"u}llermeier, E., Kanan, C., Kudithipudi, D., Lampert, C.H.,
  Mundt, M., Pascanu, R., Popescu, A., Tolias, A.S., van~de Weijer, J., Liu,
  B., Lomonaco, V., Tuytelaars, T., van~de Ven, G.M.: Continual learning:
  Applications and the road forward. Transactions on Machine Learning Research
  (2024), \url{https://openreview.net/forum?id=axBIMcGZn9}

\bibitem{wang2021afec}
Wang, L., Zhang, M., Jia, Z., Li, Q., Bao, C., Ma, K., Zhu, J., Zhong, Y.:
  Afec: Active forgetting of negative transfer in continual learning. Advances
  in Neural Information Processing Systems  \textbf{34},  22379--22391 (2021)

\bibitem{wang2024CSoCL}
Wang, L., Zhang, X., Su, H., Zhu, J.: A comprehensive survey of continual
  learning: Theory, method and application. IEEE Transactions on Pattern
  Analysis and Machine Intelligence  \textbf{46}(8),  5362--5383 (2024).
  \doi{10.1109/TPAMI.2024.3367329}

\bibitem{Wang2022dualprompt}
Wang, Z., Zhang, Z., Ebrahimi, S., Sun, R., Zhang, H., Lee, C.Y., Ren, X., Su,
  G., Perot, V., Dy, J., Pfister, T.: Dualprompt: Complementary prompting for
  rehearsal-free continual learning. In: Proceedings of the European Conference
  on Computer Vision (ECCV). p. 631–648. Springer-Verlag, Berlin, Heidelberg
  (2022). \doi{10.1007/978-3-031-19809-0\_36}

\bibitem{zenke2017SI}
Zenke, F., Poole, B., Ganguli, S.: Continual learning through synaptic
  intelligence. In: International conference on machine learning. pp.
  3987--3995. PMLR (2017)

\bibitem{gao2024prompts}
Zhanxin~Gao, Jun~CEN, X.C.: Consistent prompting for rehearsal-free continual
  learning  (2024)

\bibitem{zhou2023RevisitingCIL}
Zhou, D.W., Cai, Z.W., Ye, H.J., Zhan, D.C., Liu, Z.: Revisiting
  class-incremental learning with pre-trained models: Generalizability and
  adaptivity are all you need. International Journal of Computer Vision  (2023)

\end{thebibliography}

\end{document}